# Theory of Generative Deep Learning Ⅱ:Probe Landscape of Empirical Error via Norm Based Capacity Control

Wendi Xu[1,3], Ming Zhang[1,2]

[1]Xinjiang Astronomical Observatories, Chinese Academy of Sciences, Urumqi 830011, China
[2]Key Laboratory for Radio Astronomy, Chinese Academy of Sciences, Nanjing 210008, China
[3]University of Chinese Academy of Sciences, Beijing 100876, China
xuwendi@xao.ac.cn, zhang.ming@xao.ac.cn

**Abstract:** Despite its remarkable empirical success as a highly competitive branch of artificial intelligence, deep learning is often blamed for its widely known low interpretation and lack of firm and rigorous mathematical foundation. However, most theoretical endeavor is devoted in discriminative deep learning case, whose complementary part is generative deep learning. To the best of our knowledge, we firstly highlight landscape of empirical error in generative case to complete the full picture through exquisite design of image super resolution under norm based capacity control. Our theoretical advance in interpretation of the training dynamic is achieved from both mathematical and biological sides.

## 1 Introduction

### 1.1 Discriminative V.S. generative deep learning: landscapes of empirical error

Despite its remarkable empirical success as a highly competitive branch of artificial intelligence, deep learning is often blamed for its widely known low interpretation and lack of firm and rigorous mathematical foundation. In [1] "Theory of deep learning Ⅱ: landscape of the empirical risk in deep learning", authors from MIT explore the landscape of empirical error in deep learning via case study of a 6-layer convolutional network (CNN) for image classification to tackle the problem of theoretical puzzles. Their works are constrained in discriminative deep learning case.

However, the full picture of deep learning is made up of both discriminative and generative cases. The former paradigm works as follow: the input data is processed into representations, then the representations will be sent to classifiers (sometimes a softmax layer), the pipelines output the desired label. Image detection, image recognition and image classification etc. fall in the regime. Whereas, the latter paradigm performs as follow: the later stages of generative case involve representations learned, then the representations are transformed to "generate"(where the name of "generative" deep learning comes from) new data. Decoders in machine translation and image caption "generate" the target language from codes, decoders in image denoising and image super resolution also "generate" the image from its representations.

In order to complete the full picture, we make a step towards the scarcely explored landscape of empirical error in generative deep learning. Our theoretical and experimental explorations are conveyed by an 8-layer CNN for image super resolution within early stages. Both discriminative(authors from MIT) deep learning and generative(we) deep learning case are shown in Table Ⅰ.

**Table Ⅰ** Comparison of landscapes of empirical error

|  | Discri. DL | Gener. DL |
|---|---|---|
| **Network** | 6-layer CNN | 8-layer CNN |
| **Task(Image)** | Classification | Super Resolution |
| **Landscape** | Empirical Error ||

### 1.2 Regularization terms: working as norm based capacity control to probe landscape of empirical error

Regularization terms accompanying the loss function will modify the weights, thus control the parameter space, control the model capacity (the number of parameters).

In this article, we choose specified norm based regularization terms, which are widely known as L1 or L2 norm weight decay, working as capacity control to probe landscape of empirical error in generative deep learning case.

### 1.3 Related works: family of regularization techniques

In practice, training big networks with regularization techniques gives superior test performance to smaller networks trained without regularization.

Family of regularization techniques includes early stopping, stochastic gradient descent (implicit regularization controlled by the number of iterations), drop out in activation of fully connected layers, DroConnect [2] in weights of fully connected layers and regularization terms of norms like spectral norm, group norm, path norm, Fish-Rao norm [3], and the probably simplest norm, L1 and L2 weight decay, which is the topic of the article. We believe that simple is not simple at first glance.

## 2 Main contributions

1. To the best of our knowledge, we first investigate the theory of generative deep learning, which is complementary to the intensively studied discriminative deep learning.

2. We design three training settings of norm based capacity control, and characterize the training dynamic of 8-layer CNN within 45 epochs to reveal the landscape of empirical error.

3. We relate mathematics and neuron science to interpret the training dynamic, towards a better understanding of theory of generative deep learning.

## 3 Theoretical preparation

### 3.1 Optimization problem of deep learning

Mathematical theory of deep learning is a combination of statistics, information theory, theory of algorithms, probability and functional analysis (Beyond math, theory of deep learning is also rooted in biology and physics.). Optimization lays at the heart of the mathematical theory [4].

The general problem of deep learning can be cast as searching for the hypothesis h in function space H, which characterize the structure of the data space of $X \times Y (X \subseteq R^P, Y \subseteq R^K)$, i.e., $h_\theta : X \to Y$, that solves the following optimization problem:

$$\min_{h \in H} L_Q(h) = \mathop{E}_{Q}(L(h(x), y)) \quad (1)$$

where Q is a probability measure over the space $X \times Y$. And h can be realized [3] by feed forward neural network of depth L with coordinate-wise activation functions $\sigma_l$: $R \to R$, i.e.,

$h_\theta(x) = \sigma_{L+1}(\sigma_L(\ldots \sigma_2(\sigma_1(x^T W^0) W^1) W^2) \ldots) W^L)$,
where the parameter vector $\theta \in \Theta_L \subseteq R^d$,
$$d = p\, k_1 + \sum_{i=1}^{L-1} k_i\, k_{i+1} + k_L K,$$
and
$\Theta_L = \{W^0 \in R^{P \times k_1}, W^1 \in R^{k_1 \times k_2}, \ldots, W^L \in R^{k_L \times K}\} \quad (2)$

We set all bias terms to zero for simplicity, and set $\sigma(z) = \max\{0, z\}$, i.e., ReLU.

### 3.2 Overparametrization

The most successful [1] deep CNNs such as VGG and ResNets are best used with a degree of "overparametrization", that is more weights than data points. Those practical and theoretical results are valid in the discriminative deep learning case.

Without rigorous proof, we extend the results to generative deep learning case, and assume that a proper overparametrization also works well for deep learning based image super resolution.

### 3.3 Duration of landscape: 45 epochs

Why 45? SRCNN [5] is trained for more than $10^9$ steps, which is not computationally feasible. We need to reduce the number of steps while still attaining comparable performance.

We design a test model of 7-layer CNN, WARSHIP-XZNet [6] with 936778 parameters, for image super resolution to gain peak signal to noise ratio (PSNR) of 36.58, by 45 epochs, 6601 steps using [7] 91 images.

So we also set number of our epochs to be 45, number of steps to be 6601steps.

## 4 Empirical settings and result

### 4.1 Model setting：945318, WARSHIP-XZNetβ

Image super resolution is the process of inferring a High-Resolution (HR) version of a Low-Resolution (LR) input image, which could be divided into 3 subtasks, corresponding to 3 subnets.

The subtask of embedding sub-net (Enet) is the transformation from image space to representation space. Whereas the function of reconstruction sub-net (Rnet) is converting representation back to image. The bridge between those two subnets is the inference sub-net (Inet), which will compute the map from low resolution representation (output of Enet) to high resolution representation (input of Rnet).

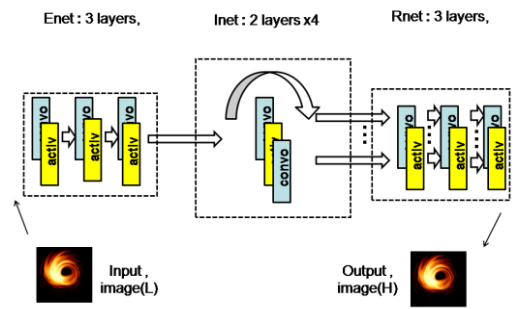

**Figure 1** overall architecture of WARSHIP-XZNetβ

Notation: "Conv layer" mentioned below means a convolutional layer and its following activation layer. In Enet, the 1st Conv layer stores the input/image, then the 2nd Conv layer transforms the image to features, finally the 3rd shrinking Conv layer with filter size of 1x1 combine 2N input features into N out features. In Inet, residual block with 1 Conv layer and additional convolutional layer is recurred 4 times. We extract every intermediate output of features by each recurrence to Rnet. In Rnet, expanding Conv layer with filter size of 1x1 inverses the sparsity in Inet, then another Conv layer transform features into several images, finally a weighed Conv layer weights the images into a single image. The Figure1 is the architecture of WARSHIP-XZNetβ. (The copyright of two pictures of accretion disk of super massive black hole in figure 1 belongs to Hotaka Shiokawa .)

Our model WARSHIP-XZNetβ is fully CNN of 3 + 2 + 3 = 8 layers. We set ReLU as activation layers and 192 features as input of 2nd Conv layer. In each Conv layer, filter size is 3x3 except that in shrinking and expanding layers with 1x1.The total parameters is 945 318. 945318 >936778 in WARSHIP-XZNet, satisfying the requirement of overparametrization.

### 4.2 Training settings : norm based capacity control

1.Empirical Error

The mean squared error of final output or empirical error is

$$L_1(\theta) = \sum_{i=1}^{N} \frac{1}{2N} \| y^{(i)} - y^{(i)}_{learned} \| \quad (3)$$

where θ is the overall parameter set of the model, N is number of training samples; $y^{(i)}$ is the i-th training sample, $y^{(i)}_{learned}$ is the output image learned by the model.

2.Regularization Terms Work as Norm Based Capacity Control

For intermediate outputs, the cost function is

$$L_2(\theta) = \sum_{i=1}^{N} \sum_{r=1}^{R} \frac{1}{2RN} \| y^{(i)} - y_r^{(i)} \|$$
$$y^{(i)}_{learned} = \sum_{r=1}^{R} w_r y_r^{(i)}$$

where N, $y^{(i)}$ and $y^{(i)}_{learned}$ are the same as (3), R is the times of recurrence, $w_r$ denotes the weight that will average images extracting from 4 recurrences into the final high resolution image, $y_r^{(i)}$ is the output image extracting from the r-th recurrence.

We devise 3 settings of regularization terms (weight decay) working as different norm based capacity control. Notation: $\theta(E)$, $\theta(I)$, and $\theta(R)$ are the parameter space of Enet, Inet and Rnet, respectively, $\|\|\|_2$ and $\|\|\|_1$ indicate L2 and L1 norm, λ is a balancing parameter and set to 0.0002.

All-L2 means that all subnets are armed with L2 norm. That is,

$$L_3(\theta) = \lambda \| \theta(E) \|_2 + \lambda \| \theta(I) \|_2 + \lambda \| \theta(R) \|_2.$$

In setting of Mix, Enet and Rnet are imposed with L2 norm, while Inet is equipped with L1 norm. That is,

$$L_3(\theta) = \lambda \| \theta(E) \|_2 + \lambda \| \theta(I) \|_1 + \lambda \| \theta(R) \|_2.$$

For All-L1 setup, all subnets are controlled by L1 norm. That is,

$$L_3(\theta) = \lambda \| \theta(E) \|_1 + \lambda \| \theta(I) \|_1 + \lambda \| \theta(R) \|_1.$$

Thus we get the overall cost function,
$$L(\theta) = (1-\alpha)L_1(\theta) + \alpha L_2(\theta) + L_3(\theta).$$
where α is the weight of $L_2(\theta)$. The chain of 3 consecutive changes works as follow: different norm of $L_3(\theta)$ causes $L(\theta)$ to vary, then variation of $L(\theta)$ causes $W^i$(i=0,…,L) in (2) to be updated, finally update of $W^i$(i=0,…,L) causes $L_1(\theta)$, i.e., empirical error, to change. In the chain, update of $W^i$(i=0,…,L) directly control the model capacity. Details of norm based capacity control are illustrated below in Table Ⅱ.

**Table Ⅱ** Norm based capacity control(weight decay).

| Norm Based Capacity Control | | | |
|---|---|---|---|
| Training Settings | All-L2 | Mix | All-L1 |
| Network | Share the same 8-layer CNN | | |
| Norm on Enet | L2 | L2 | L1 |
| Norm on Inet | L2 | L1 | L1 |
| Norm on Rnet | L2 | L2 | L1 |

3.Training Strategy

Our training dataset is also dataset of 91 images.

We adopt the same training setting as DRCN [13]. Split training images into 41 by 41 patches with stride 21 and 64 patches, which are used as mini-batch for stochastic gradient descent. We use the method described in [8] to initialize weights in non-recursive layers, and set all weights to zero except self-connections (connection to the same neuron in the next layer) for recursive convolutions. Bias are set to zero. We choose the beginning value of learning rate to be 0.01 and then decreased it by a factor of 10 if the validation error dose not decrease for 5 epochs.

Under the theoretical guideline in 3.3 ,the duration of training dynamic is set to 45 epochs (6601 steps) to explore landscape of empirical error within early stages.

### 4.3 Result of training dynamic: landscape of empirical error

PSNR is inversely proportional to the empirical error. The higher PSNR implies lower empirical error.

The training dynamic is viewed from 3 consecutive stages, 1-15 epochs, 16-30 epochs and 31-45 epochs, which character the landscape of empirical error. The ultimate PSNR of 3 settings are near 35, which is above 33.66 gained by traditional super resolution method, Bicubic.

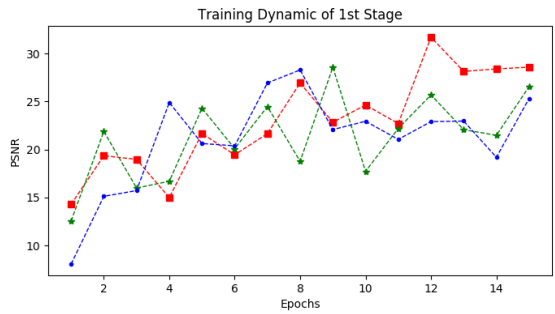

**Figure 2** Training dynamic of 1st stage. ☆ denotes setting of Mix, □ is for All-L2 and ● represents All-L1

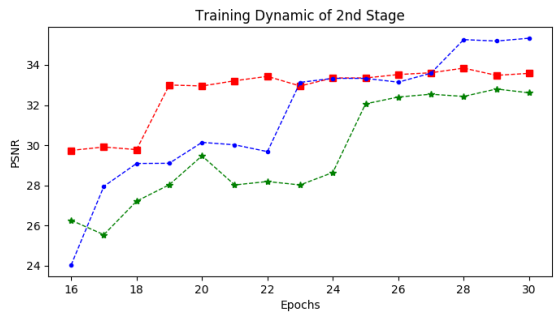

**Figure 3**　 Training dynamic of 2nd　 stage.

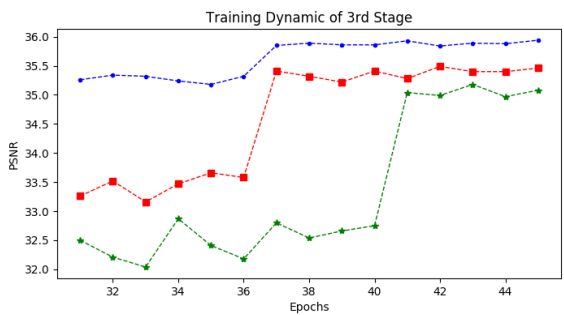

**Figure 4**　 Training dynamic of　3rd　 stage.

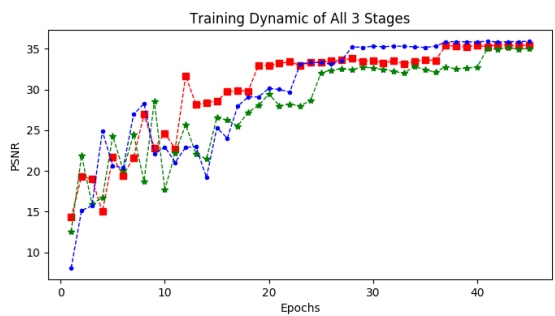

**Figure 5** Training dynamic of All 3 stages.

## 5 Discussion and conclusion: interpretation of landscapes that relates mathematics and neuroscience

### 5.1 Mathematical side of regularization terms: different sparsity

Regularizing with an L2 norm [9] is known as ridge regression in statistics and Tikhonov regularization in analysis. L2 regularization involves adding an extra term to the cost functions that penalizes the sum of squares of weight, leading to small weights, which may or may not be zero.

L1 regularization is known as Lasso in statistics, penalize the absolute values of the weights, cause many parameters exactly equal to zero, produce a sparse parameter vector. As a comparison, sparsity gained by L1 tend to be higher than that of　L2.

### 5.2 Biological side of regularization terms: different population sparseness

There are four definitions [10] of sparseness for firing patterns in research of systems neuroscience, low mean firing rate, lifetime sparseness, information per spike and population sparseness. The biological corresponding mechanism of L2 and L1 regularization tend to be population sparseness, which means that the population response distribution elicited by each stimulus is peaked. A peaked distribution contains a lot small (or 0) values and few large values. It has been suggested that neural codes with high population sparseness are advantageous because they resemble the inherently sparse structure of the sensory environment.

Based on the mathematical analysis of sparity mentioned above, L2 regularization is inclined to be higher population sparseness than L2.

### 5.3 Different subnets correspond to different visual　 areas: preservation and distillation

We tend to conjecture that Enet and Rnet is very biologically-plausible to primary visual cortex (V1). Enet and Rnet is about simple coder and decoder between images and representations. Correspondingly, early visual areas like primary visual cortex [11] encode the sensory features such as visual edges as sparse representations, which is less abstract than higher representations. Both artificial and biological encoder/decoder should preserve/retain the full/most information of the images, and involve less information distillation. To the opposite side, distillation is quite necessary for higher/later stages in image classification to perform more abstract representations biologically and artificially. As we meet at the beginning of Introduction, for the discriminative deep learning case, when deep CNNs distill irrelevant representations, they show nice/necessary properties of [12] invariance and selectivity (even need pooling layer to discard irrelevant representation), towards strong robustness and high efficiency.

Inet enables the inference, which works in a quite different mechanism from that in Enet and Rnet. It rewrite the key and core part of physical law which governs the transformation from low resolution image to high resolution image. It may share some common features with higher inference in neural system(like image classification), but to a great extent, seems to work in a distinguish style(at least, our eyes can not perform task of image super resolution.). So we should be careful to absorb the biological inspiration from V2 and V4.

### 5.4 Interpretation of landscapes of empirical error mathematically and biologically

For Enet and Rnet, comparison of Mix and All-L1 during 2nd and 3rd stages (We abandon the 1st stage which may be caused by too many factors.) tends to confirm that regularization term of L1 norm is more suitable than L2 norm to build a better landscape of empirical error. Mathematically, L1 exhibits higher sparse mechanism than L2. Additional biological observation discussed in **5.2** and **5.3** tend to prove that higher population sparseness in primary visual cortex is a better candidate for Enet and Rnet.

For Inet ,comparison of All-L2 and Mix during 2nd and 3rd stages (We also abandon the 1st stage which may be caused by too many factors.) is inclined to conclude that L2 norm is more suitable than L1 norm to achieve a better landscape of empirical error. Whereas, any biological clue/inspiration for Inet is not clear.

## 6 Theoretical exploration in generative deep learning in the future: beyond vision towards NLP

Our open theoretical insights into both discriminative and generative deep learning should benefit from interaction of 3 main AI tasks, that is, vision, audio and natural language processing(NLP).

What does matter is the essence, not the superficies. For example, image super resolution [6], image caption and machine translation [15] may sound quite different at first sight, but they actually belong to the same issue when viewed from lens of theoretical generative deep learning(at least, when considering information preservation). So in the future, beyond vision, we may experimentally and theoretically explore two NLP tasks of image caption(a special task that bridges vision and NLP) and machine translation to reveal the veil of theoretical generative deep learning, contributing to the [14]science of intelligence.

### Acknowledgements

This work was supported by the National Basic Research Program of China (2012CB821804 and 2015CB857100), the National Science Foundation of China (11103055 and 11773062) and the West Light Foundation of Chinese Academy of Sciences (RCPY201105, 2017-XBQNXZ-A-008 and 2016-QNXZ-B-25).